\documentclass[]{jair}

\setcopyright{cc}
\acmDOI{10.1613/jair.1.xxxxx}

\JAIRAE{Insert JAIR AE Name}
\JAIRTrack{} 
\acmVolume{4}
\acmArticle{6}
\acmMonth{8}
\acmYear{2026}

\RequirePackage[
  datamodel=acmdatamodel,
  style=acmauthoryear,
  backend=biber,
  giveninits=true,
  uniquename=init
  ]{biblatex}

\begin{document}

\title[Concept of Possibility]{A Concept of Possibility for Real-World Events}

\author{Daniel G. Schwartz}
\authornote{Corresponding Author.}
\orcid{0000-0002-5983-884X}
\email{dgschwartz@fsu.edu}
\affiliation{%
  \institution{Florida State University}
  \city{Tallahassee}
  \state{Florida}
  \country{USA}
}

\renewcommand{\shortauthors}{Schwartz}

\begin{abstract}
This paper offers a new concept of {\it possibility} as an alternative to the concept originally introduced by L.A. Zadeh in 1978. This new version was inspired by the original but, formally, has nothing in common with it other than that they both adopt the same interpretation of the logical connectives.  Moreover, rather than seeking to provide a general notion of possibility, this focuses specifically on the possibility of a real-world event.  An event is viewed as having prerequisites that enable its occurrence and constraints that may impede its occurrence, and the possibility of the event is computed as a function of the probabilities that the prerequisites hold and the constraints do not.  This version of possibility might appropriately be applied to problems of planning.  When there are multiple plans available for achieving a goal, this theory can be used to determine which plan is most possible, i.e., easiest or most feasible to complete.  It is speculated that this model of reasoning correctly captures normal human reasoning about plans.  The theory is elaborated and an illustrative example for vehicle route planning is provided. There is also a suggestion of potential future applications. \footnote{This work unifies and expands material formerly published as \cite{schwartz2016possibility} and \cite{schwartz2017possibility}.  The expansion includes a substantial revision of the introductory section and a new section discussing the differences between possibility and probability, and it suggests some potential future application domains.  It also provides new examples to illustrate the main ideas, and it corrects a few mathematical oversights in the previous publications.  In addition, it adds an appendix that provides the results of a simulation and a link to the code for the simulator.}
\end{abstract}



\received{20 February 2007}
\received[accepted]{5 June 2009}

\maketitle

\section{Introduction}
Possibility theory was introduced by Lotfi A. Zadeh in 1978 \cite{zadeh1978fuzzy}.  The essential idea was that, in a manner to explained in the following, a fuzzy set membership function can be interpreted as a possibility distribution.  This led to the offering of possibility theory as an alternative to probability theory as a formulation of uncertainty.  As stated by Zadeh, a possibility value can be interpreted as a measure of ``the ease'' with which an event can occur.  This contrasts with the notion of a probability value, which measures the likelihood that the event will occur.
  
That paper caught the attention of numerous researchers who proceeded to advance this theory.  Prominent among these are the French mathematicians, Didier Dubois and Henri Prade, who produced a book on this subject \cite{dubois1988possibility} and over the years have published many papers on this topic.   Related works have also appeared in an early collection edited by Ronald Yager \cite{yager1982fuzzy}, and in the more recent collection by Paul Wang \cite{wang2012advances}, both of whom have helped to advance the theory.   There are many others.   Zadeh himself returned to the topic several times, and the subject now enjoys a rich literature.  

Two more recent papers are \cite{dubois2015possibility} and \cite{dubois2016practical}.  These review most of what has been accomplished to-date and can be taken as representing the current state of the art on this topic.  The details of that approach are well-documented in the cited references and will not be repeated here.  But briefly, the essential idea proposed by Zadeh is that a fuzzy membership function can be interpreted as a possibility distribution under the following semantic reinterpretation. 

If one says that a particular person has a 0.7 degree of membership in the collection of tall people, this can be interpreted as saying that there a 0.7 {\it degree of possibility} for assigning the attribute ``tall" to that person. Accordingly, the fuzzy set membership function can be interpreted as a {\it possibility distribution}.  This notion of possibility can be extended to real-world events, where the possibility degree of the event is taken as a measure of the ``ease" with which that event may occur, or more exactly, the degree with which the attribute ``easy to occur" may be assigned to that event. 

Whereas the present paper offers a new way of looking at the concept of possibility, it in no way usurps the original. It only presents an alternative.  However, it does address an inherent weakness in that theory---this being that the notion of possibility provided in that version is purely subjective. This is because, in general, there are no objective methods for determining the values of a fuzzy membership function.  

The subjectivity of this notion of possibility is mentioned by Zadeh on p. 7 of \cite{zadeh1978fuzzy}: 

\begin{quote}

``since our intuition concerning the behaviour of possibilities is not very reliable, a great deal of empirical work would have to be done to provide us with a better understanding of the ways in which possibility distributions are manipulated by humans.  Such an understanding would be enhanced by the development of an axiomatic approach to the definition of subjective possibilities---an approach which might be in the spirit of axiomatic approaches to the definition of subjective probabilities.''

\end{quote} 

\noindent I believe that this statement can be taken as an implicit acknowledgment that the given notion of possibility is inherently subjective. 

This may be contrasted with the standard notion of probability, which affords both subjective and objective methods for determining a probability.  The objective methods comprise the methods of statistical samplings.  If one wants to know, for example, the probability that a newborn child will suffer from autism, one can perform a sampling of children and find, for example, that this occurs in 1 out of every 50 children, in which case the probability can be given as $1/50 = 0.02$. There is no similar procedure for possibility.  This, I will speculate, is the reason that, while possibility theory is now quite advanced, there have been few practical applications. 

The version of possibility being introduced here stands in contrast with the original in that, by basing a notion of possibility on the concept of probabilities for prerequisites and constraints, this can provide an objective method for computing a possibility degree.  While one still has the option of employing subjective probabilities if desired, these probabilities can be determined objectively as long as the relevant data can be obtained.  For this reason, this new concept should be more amenable to real-world applications.   

In comparison with the original version, this approach does not employ fuzzy sets.  Accordingly, while this work has been inspired by the works of Zadeh and the others, it comprises a completely different way of looking at the concept of possibility.

Having said this, however, it happens that the present system does share with fuzzy sets theory the interpretation of the logical ``not'', ``or'', and ``and'', respectively, as the arithmetic $1-$, {\it max}, and {\it min} operations.  It does this, moreover, because these connectives provide an intuitively plausible model of natural human reasoning in these contexts.  This is discussed more fully in the following Section 8.  

This interpretation of the logical connectives is by no means unique to fuzzy sets theory, however, and logics of this type have been studied extensively in the literature on multi-valued logics.  The earliest discussion of this kind of logic may attributed to Jan \L ukasiewicz, whose formulation is presented in \cite{rescher1969many}, p. 337, as the system $L_{\aleph_1}$. This logic provides the same interpretations of the logical connectives as in the present work, and it additionally provides an interpretation for the  implication connective $\to$.   Thus, the logic being used here may be viewed as a subsystem of $L_{\aleph_1}$ and, for want of a better term, will be referred to simply as the {\it multivalent logic}.

Real-world events are herein regarded as being {\it context dependent}, where the prerequisites and constraints comprise the context.  The following Section 2 offers an intuitive rationale for this approach in terms of a simple example.  Section 3 presents a precise formalization.  Section 4 continues the example presented in section 2, adding a new feature.  Section 5 expands the formalism presented in Section 3 to encompass this added feature.  A key issue addressed here is how to determine when two contexts are equivalent.  Section 6 considers the case of more complex events, where the possibility of an event depends on the possibilities of precursor events.  Section 7 presents an illustration of these ideas in a simple example involving vehicle routing via waypoint navigation.  Section 8 compares the alternative formulation using probability theory.  Section 9 suggests prospects for future applications.  Section 10 provides concluding remarks.  An appendix presents the results of a simulation of the waypoint example and provides a link to the current version of the software.

\section{Intuitive Rationale}

To illustrate the core ideas, suppose that ABC Corporation wishes to develop a new product line and needs to build a new factory as well as acquire additional employees.  Let us assume that this requires raising sufficient capital to finance the project and, in addition, requires the availability of suitable new employees.  Then ``capital'' and ``employees'' are prerequisites.  It is proposed to compute the degree of possibility for the new product line as a function of the probability that ABC Corporation will be able to obtain the necessary capital and employees, to wit, \medskip

${\rm Poss}({\rm new\_product}) = {\it min}[{\rm Prob}({\rm capital}), {\rm Prob}({\rm employees})]$ \hfill (1) \medskip

\noindent where Prob is a standard probability measure such as given by the well-known Kolmogorov axioms, \break cf. \cite{kolmogorov2018foundations}.  As noted previously, the {\it min} operation represents the logical ``and'' in the multi-valent logic.  

While the probabilities in this example might not lend themselves to evaluation by statistical samplings, it is clear that in many applications they would.  Consider for example the following weather prediction from a local newspaper \cite{burlew2015severe}:  \medskip

``Severe storms possible ahead of Christmas Day''

``There is a possibility of 2-3 inches of rain.'' \medskip

\noindent These possibilities are based on probabilistic models involving such factors as temperature, humidity, cloud formations, and wind currents.  

In this spirit, the proposed approach uses the methods of probability theory to determine the likelihood that the prerequisites will be satisfied and then uses the multivalent logic to give the degree of possibility for the event.   Note that, in this example regarding ABC Corporation, we are here considering only the possibility that they may choose to develop the new product line, or following Zadeh, the ``ease'' with which this might occur, and not the probability that they will actually do so.  This idea of interpreting a possibility degree for an event as a ``degree of ease'' with which the event may occur is discussed at some length on pages 8-9 of \cite{zadeh1978fuzzy}.

Now suppose that ABC Corporation has learned that there may be environmental issues that could interfere with their plans to build a new factory on a particular site.  This would be a constraint.  In this case, the foregoing computation becomes \medskip

${\rm Poss}({\rm new\_product}) = {\it min}[{\rm Prob}({\rm capital}), {\rm Prob}({\rm employees}),$

\hskip 9.6em ${\rm Prob}({\lnot\rm environmental\_issue})]$ \hfill (2) \medskip

\noindent or equivalently \medskip

${\rm Poss}({\rm new\_product}) = {\it min}[{\rm Prob}({\rm capital}), {\rm Prob}({\rm employees}),$

\hskip 9.6em $1-{\rm Prob}({\rm environmental\_issue})]$ \hfill (3) \medskip

\noindent using the probability theory representation of the logical ``not''.  In effect, $1-{\rm Prob}({\rm environmental\_issue})$ measures the degree to which the constraint ``environmental\_issue'' is mitigated.  Note that this amounts to regarding the mitigation of the constraint as a precondition.  In general, any constraint $c$ can be construed as specifying a   corresponding precondition having the form $\lnot c$. 

These examples illustrate an intuitively plausible means for working with conjunctions of contextual elements (prerequisites and constraints).  For a further example, involving disjunction and using the multivalent ``or'', suppose that ABC Corporation has two employees, Jack and Jill, that they believe could be effective managers of their new factory, so that the success of their new enterprise depends on at least one of these people being interested.  This can be expressed as \medskip

${\rm Poss}({\rm new\_product}) = {\it min}[{\rm Prob}({\rm capital}), {\rm Prob}({\rm employees}),$

\hskip 9.6em $1-{\rm Prob}({\rm environmental\_issue})$, 

\hskip 9.6em ${\it max}[{\rm Prob}({\rm Jack}), {\rm Prob}({\rm Jill})]$ \medskip

 But now suppose they have recently learned that Jack has health issues that might make him unavailable.  Then the formula for the event can be written as \medskip

${\rm Poss}({\rm new\_product}) = {\it min}[{\rm Prob}({\rm capital}), {\rm Prob}({\rm employees}),$

\hskip 9.6em $1-{\rm Prob}({\rm environmental\_issue})$, 

\hskip 9.6em ${\it max}[{\rm Prob}(\lnot{\rm Jack\_unavailable}), {\rm Prob}({\rm Jill})]$ \hfill (4) \medskip

\noindent or \medskip

${\rm Poss}({\rm new\_product}) = {\it min}[{\rm Prob}({\rm capital}), {\rm Prob}({\rm employees}),$

\hskip 9.6em $1-{\rm Prob}({\rm environmental\_issue})$, 

\hskip 9.6em ${\it max}[1-{\rm Prob}({\rm Jack\_unavailable}), {\rm Prob}({\rm Jill})]$ \hfill (5) \medskip

From a purely logical standpoint there is no difference between regarding a constraint as such and regarding the semantic negation of the constraint as a prerequisite, for which reason the very consideration of constraints might seem superfluous.  Such do play a role, however, as a means of making the constraints explicit in the context of the event.  For example, the negated constraint $\lnot{\rm Jack\_unavailable}$ might have been expressed as the prerequisite ${\rm Jack\_available}$, but taking the former approach emphasizes the fact that Jack's possible unavailability might be an issue. 

Since probabilities are always in the range $[0,1]$, so also will be the values returned by the foregoing applications of $1-$, {\it min} and {\it max}, and therefore also the values of the function {\rm Poss}.  These values will be referred to as {\it possibility degrees}.

\section{Formalization}

These considerations motivate the following formal definitions.  For an event $E$, any proposition $p$ can serve as a prerequisite, and any proposition $c$ can serve as a constraint.  Let us define the {\it contextual constructs} for event $E$ by:

\begin{enumerate} 
\item If $p$ is a prerequisite for $E$, then $p$ is a contextual construct for $E$.
\item If $c$ is a constraint for $E$, then $(\lnot c)$ is a contextual construct for $E$.
\item If $C_1$ and $C_2$ are contextual constructs for $E$, then ($C_1\land C_2$) is a contextual construct for $E$.
\item If $C_1$ and $C_2$ are contextual constructs for $E$, then ($C_1\lor C_2$) is a contextual construct for $E$.
\item Nothing is a contextual construct for $E$ except as required by the items 1 through 4.
\end{enumerate}

\noindent Outermost surrounding parentheses will be omitted when not required for disambiguation.  A contextual construct either of the form $p$ where $p$ is a prerequisite or of the form $\lnot c$ where $c$ is a constraint will be an {\it atomic} contextual construct.

Given an event $E$, let us define the {\it possibility valuation} $v$ for contextual constructs for $E$ by: 

\begin{enumerate}
\item If C is an atomic contextual construct for $E$, then $v(C)={\rm Prob}(C)$. 
\item If $C$ is of the form $(C_1\land C_2)$ where $C_1$ and $C_2$ are contextual constructs for $E$, then $v(C)=$\break${\it min}(v(C_1),v(C_2))$.
\item  If $C$ is of the form $(C_1\lor C_2)$ where $C_1$ and $C_2$ are contextual constructs for $E$, then $v(C)=$\break${\it max}(v(C_1),v(C_2))$.
\end{enumerate}

For the purposes of this model, it is assumed that the atomic contextual constructs are statistically independent, i.e, there is no correlation among them and the occurrence of any one of them has no effect on the occurrence of any others. 

Let us say that a contextual construct for an event $E$ is {\it complete} if it is considered to be a full description of the relevant context for $E$ in terms of the event's prerequisites and constraints.  Then if $C$ is a complete contextual construct for $E$, set \medskip

${\rm Poss}(E)=v(C)$ \medskip

Note that this notion of completeness for contextual constructs is fundamentally intuitive and inherently ambiguous inasmuch as it depends on a human user's perception of the relevant situation.   One person may consider a different set of prerequisites and constraints than another as being necessary.  This ambiguity notwithstanding, however, the notion serves a useful purpose, once a particular conceptual construct has been chosen and agreed upon.    

Let us now illustrate this formalism with a return to the foregoing examples.

\section{Examples Revisited} 

For a simple case, consider $E$ as the initial event of ABC Corporation creating a new product.  The prerequisites are $p_1={\it sufficient\ capital}$ and $p_2={\it sufficient\ employees}$ and both are required, so the complete contextual construct for $E$ is \medskip

$C=(p_1\land p_2)$ \medskip 

\noindent and the foregoing definitions give \medskip

${\rm Poss}(E)=v(C)$

\hskip 3.8em $={\it min}[v(p_1),v(p_2)]$

\hskip 3.8em $={\it min}[{\rm Prob}(p_1),{\rm Prob}(p_2)]$ \medskip
  
\noindent Thus one obtains the same result as described in the intuitive rationale (1). 

Next consider the slightly more complex event in (2) and (3).  Let $c={\it environmental\_issue}$.  Then the complete contextual construct becomes \medskip

$C=(p_1\land p_2\land\lnot c)$ \medskip 

\noindent which evaluates as \medskip

${\rm Poss}(E)=v(C)$

\hskip 3.8em $={\it min}[v(p_1),v(p_2),v(\lnot c)]$

\hskip 3.8em $={\it min}[{\rm Prob}(p_1),{\rm Prob}(p_2), {\rm Prob}(\lnot c)]$
 
\hskip 3.8em $={\it min}[{\rm Prob}(p_1),{\rm Prob}(p_2), 1- {\rm Prob}(c)]$  \medskip  

As a further example, consider the final, more complex case illustrated in (4) and (5).  Again, call the event $E$.  Here let $p_1$ and $p_2$ be as above, let $c_1={\it environmental\_issue}$, let $c_2={\it Jack\_unavailable}$, and let $p_3={\rm Jill}$.  Then a complete contextual construct for $E$ is \medskip

$C=p_1\land p_2\land\lnot c_1\land(\lnot c_2\lor p_3)$ \hfill (6) \medskip

\noindent which evaluates as \medskip

${\rm Poss}(E)=v(C)$

\hskip 3.8em $={\it min}[v(p_1),v(p_2),v(\lnot c_1),max[v(\lnot c_2),v(p_3)]$

\hskip 3.8em $={\it min}[{\rm Prob}(p_1),{\rm Prob}(p_2), {\rm Prob}(\lnot c_1),$

\hskip 6em${\it max}[{\rm Prob}(\lnot c_2),{\rm Prob}(p_3)]$ 

\hskip 3.8em $={\it min}[{\rm Prob}(p_1),{\rm Prob}(p_2), 1-{\rm Prob}(c_1),$

\hskip 6em ${\it max}[1-{\rm Prob}(c_2),{\rm Prob}(p_3)]$ \medskip
 
Note that there can be more than one complete contextual construct depending on the manner in which these are built up from atomic constructs.  For example, \medskip

$C'=(p_3\lor\lnot c_2)\land\lnot c_1\land p_2\land p_1$ \hfill (7) \medskip

\noindent can also serve as a complete contextual construct for the given event.  

\section{Formalization Continued}

Let us henceforth refer to a complete contextual construct for an event simply as a {\it context} for the event.  Note that a context for an event is taken to include not only the prerequisites and constraints for the event, but also the manner in which these are viewed as being interrelated.  Given that there can be more than one complete context for the same event, the question arises whether all such contexts will evaluate to the same possibility degree.  

It might be reasonable to expect that such contexts should be logically equivalent when thus regarded as propositions of the classical propositional calculus (PC).  It happens, however, that this in itself is not sufficient to guarantee that they will evaluate to the same possibility degree.  The reason is that it is not generally true that, if two propositions of PC are equivalent with respect to that logic, they will also be equivalent when interpreted in the multivalent logic. This was pointed out to me in some exchanges on the Berkeley Initiative on Soft Computing (BISC) email list with Vladik Kreinovich and Dana Scott in reference to the logic of fuzzy sets, which happens to be identical to the multivalent logic considered here.  Kreinovich notes that the propositions $p\lor\lnot p$ and $q\lor\lnot q$ are equivalent in classical logic because they are both tautologies, but if $p$ and $q$ are assigned different multivalent truth values then the two propositions typically will evaluate differently, e.g., setting $v(p)=0.4$ and $v(q)=0.8$ give $v(p\lor\lnot p)=0.6$ and $v(q\lor\lnot q)=0.8$.  Scott's example consists of the propositions $(p\lor\lnot p)\land q$ and $(p\land\lnot p)\lor q$, which are equivalent in classical logic but evaluate to 0 and 0.5 respectively in the multivalent logic if one sets $v(p)=0.5$ and $v(q)=0$. This accordingly raises the question of what additional conditions on the constructs might ensure that they evaluate to the same possibility degree.  The following proposes an answer. 
 
To begin, let us review some well-known facts regarding PC.  The language of PC uses {\it propositional variables} ${\bf p}_1,{\bf p}_2,\ldots$, some {\it connectives} denoted $\land$, $\lor$, and $\lnot$, and parentheses $($ and $)$.  The {\it propositions} are defined by: \medskip

\begin{enumerate}
\item propositional variables are propositions,
\item if $p$ and $q$ are propositions, then so are $(\lnot p)$, $(p\land q)$, and $(p\lor q)$,
\item nothing is a proposition except as required by items 1 and 2.
\end{enumerate}
 
The three given connectives are adequate for the language of PC, since the connectives $\to$ and $\leftrightarrow$ can be defined in terms of these by taking $(p\to q)$ as an abbreviation for $(\lnot p\lor q)$ and taking $(p\leftrightarrow q)$ as an abbreviation for $((p\to q)\land(q\to p))$ (see \cite{mendelson2015introduction} or \cite{hamilton1988logic}).   It is assumed that the reader is familiar with the basics of PC.  Parentheses will be dropped when the intended grouping is clear; in particular, outermost parentheses are always dropped; {\it negation}, $\lnot$, takes priority over both {\it conjunction}, $\land$, and {\it disjunction}, $\lor$; associativity is assumed to be to the right.

Taking 1 and 0 to stand for the truth values {\bf T} and {\bf F}, the classical semantics for this language can be defined in terms of a {\it valuation mapping} $v:{\rm propositions}\to\{1,0\}$ according to:  

\begin{enumerate}
\item if $p$ is propositional variable, then $v(p)\in\{1,0\}$,
\item if $p$ is of the form $\lnot q$, where $q$ is a proposition, then $v(p)=1-v(q)$,
\item if $p$ is of the form $q\land r$, where $q$ and $r$ are propositions, then $v(p)={\it min}(v(q),v(r))$,
\item if $p$ is of the form $q\lor r$, where $q$ and $r$ are propositions, then $v(p)={\it max}(v(q),v(r))$.
\end{enumerate}

\noindent The semantics for possibilistic logic can be obtained from this by taking the unit interval $[0,1]$ in place of the set $\{1,0\}$.  In this case, the members of $[0,1]$ are interpreted as {\it possibility degrees} in essentially the same sense as this term was used in the foregoing.  Two propositions $p$ and $q$ are {\it equivalent} with respect to either semantics if one has $v(p)=v(q)$ for all choices for the respective versions of $v$, i.e., all mappings into $\{1,0\}$ for classical logic, and all mappings into $[0,1]$ for possibilistic logic. 

A {\it literal} is a proposition of the form $p$ or $\lnot p$, where $p$ is a propositional variable.  A {\it disjunctive normal form} (DNF) is typically defined by saying that it  is a disjunction of one or more conjunctions, each of which is a conjunction of one or more literals (see \cite{mendelson2015introduction}, p. 23), or more exactly, that it is a proposition of the form $\bigvee_{i=1}^m(\bigwedge_{j=1}^n Q_{i,j})$, where each $Q_{i,j}$ is a literal (see \cite{hamilton1988logic}, p. 17).\footnote{The previous work \cite{schwartz2016possibility} employed conjunctive normal forms.  For the present purposes, the two approaches are equivalent.  The DNF approach is adopted here so that one may rely on the references cited in the following for the conversion algorithm {\it conv}.}  It should be noted, however, that such an expression strictly speaking does not qualify as a proposition of PC unless one assumes association to the left or right.  For example, where $p_1,p_2,p_3$ are propositions, the expressions $((p_1\land p_2)\land p_3))$ and $(p_1\land(p_2\land p_3))$ are propositions, but the expression $(p_1\land p_2\land p_3)$ is not.  Accordingly, the typical definition is in this respect ambiguous.  More exactly, what is wanted is that a proposition is a DNF if it is of the form  $\bigvee_{i=1}^m(\bigwedge_{j=1}^n Q_{i,j})$, where each $Q_{i,j}$ is a literal, and, in addition, either this expression employs associativity to the right, or it can be converted into an expression that employs associativity to the right by applying the rules of associativity, namely, for arbitrary propositions $p,q,r$,\medskip

replace $(p\lor q)\lor r)$ with $(p\lor(q\lor r))$,

replace $(p\lor(q\lor r))$ with $((p\lor q)\lor r)$,

replace $(p\land q)\land r)$ with $(p\land(q\land r))$,

replace $(p\land(q\land r))$ with $((p\land q)\land r)$. \medskip

\noindent These rules are justified by the fact that they amount to replacing a proposition with one that is logically equivalent.  Following \cite{hamilton1988logic}, p. 16, the conjunctions in a DNF will be referred to as {\it basic conjunctions}. 

It is well-known that every proposition of PC has an equivalent DNF.  Semantic arguments proving this may be found in both of the cited references.  There also is a well-known algorithm for converting a proposition into a DNF; see the page at \cite{wikipedia2026axioms}, which, in turn, references \cite{dershowitz1990rewrite}.   This algorithm may be given in the form of a recursive procedure {\it conv} as follows.  For an arbitrary proposition $\pi$ of PC, \medskip

\begin{enumerate}
\item If $\pi$ is a literal, stop.
\item If $\pi$ is of the form $\lnot\lnot p$, replace $\pi$ with ${\it conv}(p)$.
\item If $\pi$ is of the form $\lnot(p\lor q)$  replace $\pi$ with ${\it conv}(\lnot p)\land{\it conv}(\lnot q)$.
\item If $\pi$ is of the form $\lnot(p\land q)$, replace $\pi$ with ${\it conv}(\lnot p)\lor{\it conv}(\lnot q)$.
\item If $\pi$ is of the form $p\lor(q\land r)$, replace $\pi$ with $({\it conv}(p)\lor{\it conv}(q))\land({\it conv}(p)\lor{\it conv}(r))$.
\item If $\pi$ is of the form $p\land(q\lor r)$, replace $\pi$ with $({\it conv}(p)\land{\it conv}(q))\lor({\it conv}(p)\land{\it conv}(r))$.
\item If $\pi$ is of the form $(p\land q)\lor r$, replace $\pi$ with $({\it conv}(p)\lor{\it conv}(r))\land({\it conv}(q)\lor{\it conv}(r))$.
\item If $\pi$ is of the form $(p\lor q)\land r$, replace $\pi$ with $({\it conv}(p)\land{\it conv}(r))\lor({\it conv}(q)\land{\it conv}(r))$.
\end{enumerate} 

Item 2 applies the Idempotence Law, items 3 and 4 apply De Morgan's Laws, and items 5 and 6 apply the Distributive Laws.  The fact that this conversion process produces an equivalent proposition is established by the fact that these laws are all valid in PC.  It is also known that PC validates the Commutative and Associative Laws.

Let us say that two propositions $p$ and $q$ of PC are {\it strongly equivalent} if ${\it conv}(p)$ and ${\it conv}(q)$ can be converted into one another using only the Commutative and Associative Laws.  In effect, this says that, not only are $p$ and $q$ logically equivalent, but their DNFs differ from one another essentially only in the order of their basic conjunctions and the order of the literals within the basic conjunctions.  Similarly let us say that two contextual constructs $C$ and $C'$, are {\it strongly equivalent} if they are strongly equivalent when considered as propositions of PC.  

Next let us note the following facts about the multivalent logic.  First, the Distributive Laws hold.  Consider \medskip

$p\lor(q\land r)$ and $(p\lor q)\land(p\lor r)$ \medskip  

\noindent By examining the six possible arrangements of $p$, $q$, and $r$ with respect to their numerical values (i.e., $v(p)\le v(q)\le v(r)$, $v(p)\le v(r)\le v(q)$, $v(q)\le v(p)\le v(r)$, $v(q)\le v(r)\le v(p)$, $v(r)\le v(p)\le v(q)$, and $v(r)\le v(q)\le v(p)$) one can see that in all cases both of the above formulas evaluate to the same number.  Similarly for \medskip

$p\land(q\lor r)$ and $(p\land q)\lor(p\land r)$ \medskip

Second, the Commutative Law holds. To wit, for all propositions $p,q$ with possibility degrees $v(p), v(q)$, we have \medskip

$v(p\land q) = {\it min}(v(p),v(q))$

\hskip 3.7em $= {\it min}(v(q),v(p))$

\hskip 3.7em $= v(q\land p)$ \medskip

Last, the Associative Laws hold.   For all propositions $p,q,r$ with possibility degrees $v(p), v(q, v(r))$, we have \medskip

$v((p\land q)\land r) = {\it min}({\it min}(v(p),v(q)),v(r))$

\hskip 6.1em $= {\it min}(v(p), {\it min}(v(q),v(r)))$

\hskip 6.1em $=v(p\land (q\land r))$ \medskip

\noindent and similarly for $\lor$ with {\it max}. 

We are now in a position to establish the following result.  \medskip

{\bf Theorem.}  If two contextual constructs $C$ and $C'$ are strongly equivalent, then $v(C)=v(C')$.\footnote{A generalization of this result that applies to the full logic of fuzzy propositions, and not just conceptual constructs, has been published as \cite{schwartz2024strong}.  The following proof, as well as the foregoing definitions and analysis, repeats some of that material and is provided here so that the present treatment will be self-contained.} \medskip 

{\bf Proof.}  Let $C$ and $C'$ be two strongly equivalent contextual constructs.  Consider the DNFs ${\it conv}(C)$ and ${\it conv}(C')$.  Note that, because of the rules for building contextual constructs, none of items 2, 3, or 4 in the foregoing algorithm {\it conv} will need to be applied, since the types of propositions to which they are applied are not conceptual constructs.  Thus, only items 5, and 6 will be employed by the conversion process, and these apply only the Distributive Laws, which we have just seen are valid in the multivalent logic.  Thus $v({\it conv}(C))=v(C))$ and $v({\it conv}(C'))=v(C'))$. 

By the definition of strongly equivalent for contextual constructs, each of ${\it conv}(C)$ and ${\it conv}(C')$ can be converted into the other using only the Commutative and Associative Laws.  Since, as shown in the foregoing, these laws are also valid in possibilistic logic, it follows that $v({\it conv}(C))=v({\it conv}(C'))$.  This together with the foregoing gives the desired result:  \medskip

$v(C)=v({\it conv}(C))=v({\it conv}(C'))=v(C')$ \medskip

\noindent $\blacksquare$ \medskip

This theorem thus establishes the desired condition for equivalence between contexts for any given event.  To illustrate, we can see that the foregoing constructs (6) and (7) are strongly equivalent, and therefore evaluate to the same possibility value.  First, let us write these propositions with the parentheses implied by assuming associativity to the right (but omitting outermost parentheses), \medskip

$C=p_1\land (p_2\land(\lnot c_1\land(\lnot c_2\lor p_3)))$ \medskip

$C'=(p_3\lor\lnot c_2)\land(\lnot c_1\land( p_2\land p_1))$  \medskip

\noindent These have the property that ${\it conv}(C)$ and ${\it conv}(C')$ can be converted into one another by using only the Commutative and Associative Laws.   Following is an application of {\it conv} to $C$:

$p_1\land (p_2\land((\lnot c_1\land\lnot c_2)\lor(\lnot c_1\land p_3)))$, using {\it conv} item (6)

$p_1\land ((p_2\land(\lnot c_1\land\lnot c_2))\lor(p_2\land(\lnot c_1\land p_3)))$, using {\it conv} item (6) 

$(p_1\land (p_2\land(\lnot c_1\land\lnot c_2)))\lor(p_1\land(p_2\land(\lnot c_1\land p_3)))$, using {\it conv} item (6) \hfill (8) \medskip

\noindent Next is an application of {\it conv} to $C'$: \medskip

$(p_3\land(\lnot c_1\land( p_2\land p_1)))\lor(\lnot c_2\land(\lnot c_1\land( p_2\land p_1)))$, using {\it conv} item (8) \hfill (9) \medskip

\noindent It is easy to see that (8) and (9) satisfy the conversion requirement. 
    
\section{Complex Events}

The foregoing has defined a notion of possibility for an event whose occurrence is predicated on a set of prerequisites and/or constraints.  This in turn gives rise to the issue of an event whose possibility is predicated on the possibilities of some precursor events.  More exactly, we here consider an event $E$ whose possibility depends on the possibility of some Boolean combination $E'$ of precursor events $E_1,\ldots,E_n$, where it is assumed that possibility values for $E_1,\ldots,E_n$ are known.  The former type of event, $E$, may be referred to as a {\it simple event} and the latter, $E'$, as a {\it complex event}.  At issue is how to compute ${\rm Poss}(E)$ given ${\rm Poss}(E_1),\ldots,{\rm Poss}(E_n)$.  

It is natural to compute the possibility of the Boolean combination $E'$ in terms of the possibilities of $E_1,\ldots,E_n$ using the standard possibility theory interpretations of $\land$, $\lor$, and $\lnot$ as the operations {\it min}, {\it max}, and $1-$ as discussed in the foregoing. The question, then, is how to determine the possibility of $E$ given the possibility of the Boolean combination $E'$.

For this it is important to note that the connection between $E$ and $E'$ is a logical relationship.  This may be contrasted with probability theory where the conditional probability of $E$ given $E'$ may be interpreted as a causality relation.  In possibility theory there is no role for the notion of causation.       The possibility relation between $E$ and $E'$ is simply that, if $E'$ is possible, then so is $E$; i.e., it is a purely logical inference.

Because of this it is natural to compute the possibility of $E$ in terms of the possibility of $E'$ by means of some possibilistic version of the classical rule of Modus Ponens.  This amounts to selecting a suitable inference operator, i.e., a functional representation of the logical $\to$.  In the present context it is reasonable to assume that the inference $E'\to E$ is true, in which case its truth value in a multivalent logic would be 1.  Given this, then a value for ${\rm Poss}(E)$ can be computed from the value for ${\rm Poss}(E')$ according to the function specified for $\to$.

 To illustrate, consider \L ukasiewicz $\to$ operator from the aforementioned \L$_{\aleph_1}$, cf. \cite{rescher1969many}, p. 337, defined by \medskip 
 
$v(E'\to E)= {\it min}(1, 1-{\rm Poss}(E')+{\rm Poss}(E))$ \medskip

\noindent In this case, if $v(E\to E') =1$, then $1-{\rm Poss}(E')+{\rm Poss(E)} \ge 1$, so that ${\rm Poss}(E) \ge{\rm Poss}(E')$.  Thus one obtains a range of values for ${\rm Poss}(E)$. Under the current circumstances, however, there is no loss in simply assuming that ${\rm Poss}(E)={\rm Poss}(E')$.  

But this is only one of many available choices for an inference operator.  Numerous other such operators have been discussed throughout the literature.  In particular nine different versions of Modus Pones are discussed in \cite{dubois1988possibility} (pp. 130--131) and ten are discussed in \cite{schwartz1989fuzzy}.  Which one to employ depends on the preferences of the user and the needs of the application. 

Note that, with the \L ukasiewicz operator under the foregoing assumptions, the degree of possibility propagates unchanged through chains of inferences.  With other versions of logical inference these degrees may degrade as the chain grows longer.  In some cases, the possibility value can degrade rather quickly toward 0.5, indicating complete uncertainty.  This may be suitable for modelling some situations, and not for others.   

\section {An Example: Vehicle Waypoint Navigation}

These ideas may be illustrated with a hypothetical real-world application.  As mentioned, possibility theory may profitably be applied to problems of planning. A plan consists of a sequence of events leading from a start state to a goal state.  Possibility theory can play a role when there is more than one plan for the same start-goal mission. The theory can be applied to compute a degree of possibility for each plan, from which one can choose the plan that has the highest possibility degree and thus may be regarded as the easiest to complete.  

Applying the theory requires that, for each event, one assigns suitable prerequisites and constraints, and then for each of these prerequisites and constraints finds data sources that can be used to estimate their probabilities.  However, if desired, or in case relevant data is not available, such probabilities can be assigned subjectively.  Given that this has been accomplished, the degrees of possibility for the various plans can be computed in the manner that is illustrated in this example.

Consider the task of navigating a vehicle (with or without a driver) through a network of city streets.  Suppose that, as depicted in Figure 1, it is desired to travel from point A to point H, and some mapping service has identified several alternative routes, also shown in Figure 1.  In this case each route comprises a plan, where the tasks consist of navigating the legs of the route, i.e., each leg comprises a task.
 Each node in the graph is a {\it waypoint} and each link between two waypoints is a {\it leg}. n this case each route comprises a plan, where the tasks consist of navigating the legs of the route, i.e., each leg comprises a task.  As can be seen in Figure 1, there are three plans, route A, B, D, G, H, route A, B, E, G, H, and route A, C, F, G, H.

 Further suppose that this is taking place in a ``smart city'', which provides the vehicle with real-time information concerning traffic conditions on all the indicated legs of the journey.  The objective is to determine the degree of possibility that the vehicle can reach waypoint H at a speed of at least the designated speed limits for the various legs. 

\begin{figure}[t]
\centerline{\includegraphics[width=2in]{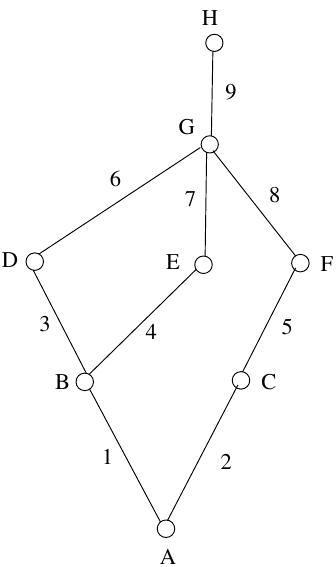}}
\caption{Example street network.}
\end{figure}
 
For the purposes of this example, assume also that the vehicle reassesses this possibility at each waypoint where it must make a decision regarding which way to proceed.  Thus, at point A the vehicle may decide to proceed to either B or C.  If it chooses B, then, after arriving at B, it must choose between D and E and then proceed to G and H.  If it chooses C, then it is committed to the path A, C, F, G, H.
 
For simplicity, let us assume that the prerequisites and constraints are the same for all legs.  In a more realistic example, of course, this would not be the case.  Let the prerequisites for all legs be $p_1=$ ``the vehicle is in proper operating condition'' and $p_2=$ ``the driver (human or robot) is competent''.  Constraints that may apply to each leg are possible causes of traffic congestion that impede the vehicle's progress.  These may include $c_1=$ ``high traffic volume (rush hours)'', $c_2=$ ``bad weather (rain, snow, ice)'', $c_3=$ ``traffic accidents'', and $c_4=$ ``road construction''. 

One might reasonably set the probability of $p_1$ to be strictly less than 1 to account for the possibility that the vehicle might suffer mechanical problems (flat tire, engine trouble).  Similarly, one might set the probability of $p_2$ to be less than 1 to account for the possibility that the driver might be in training, or impaired (alcohol, drugs), sleepy, or experience health issues.  Also, given an adequate monitoring system, these values could change during the course of the journey.  Constraints also may vary throughout the trip, e.g., a traffic accident can suddenly occur, or the daily rush hour can come into effect, or the weather can change.  While some of these probabilities may be assigned subjectively, others might be based on statistical samplings.  For example, data regarding traffic congestion during different times of the day could be used to determine the degree to which this can become a constraint.  Similar remarks may apply to the frequency of traffic accidents on a particular leg of the vehicle's journey. In addition, weather forecasts could be invoked to determine the likelihood of adverse conditions on later legs of the journey.

For the sake of this example, assume that a complete contextual construct for each leg is $C=p_1\land p_2\land\lnot c_1\land\lnot c_2\land\lnot c_3\land\lnot c_4$.  To identify these items for each leg $i=1,\ldots, 9$ shown in Figure 1, add the subscript $i$, i.e., write $C_i=p_{1,i}\land p_{2,i}\land\lnot c_{1,i}\land\lnot c_{2,i}\land\lnot c_{3,i}\land\lnot c_{4,i}$.  Then, in accordance with the foregoing theory, for each $i$, if $E_i$ is the event of the car traveling leg $i$ at the designated speed limit, it follows that \medskip

${\rm Poss}(E_i)=v(C_i)$

\hskip 4em $={\it min}({\rm Prob}(p_{1,i}),{\rm Prob}(p_{2,i}),1-{\rm Prob}(c_{1,i}),$

\hskip 5 em $1-{\rm Prob}(c_{2,i}),1-{\rm Prob}(c_{3,i}),1-{\rm Prob}(c_{4,i}))$ \medskip

The indicated probabilities may be provided by subjective or objective analysis, where by ``objective'' is here meant ``based on statistical sampling''.  For example, consider constraint $c_1$.   Traffic volume may be taken as a linguistic variable with possible values {\it high, medium, low}, where volume is measured as the number of vehicles passing a given point per minute.   For each hour of the day, the probability that the traffic will be high during that time can be determined by historical statistical samplings; e.g., the probability that congestion will be high around 5:30 PM on a Monday might be 0.9.  Other probabilities may depend on subjective evaluations or current reports.  For example, the probability of a traffic accident (constraint $c_3$) on a given leg based on statistical samplings might be generally low, say 0.1, but if the smart city system announces that there is an accident on some leg, then the evaluation immediately jumps to 1.0.  In this manner, one can determine ${\rm Poss}(E_i)$ for each leg $i$ for any time.

Accordingly, the value ${\rm Poss}(E_i)$ can be computed dynamically while the vehicle is travelling for any time of day and any day of the week.   Given this ability, the procedure for the vehicle's decisions regarding which path to take goes as follows.  

While at waypoint A, the choice is whether to proceed to B or C.  Let $E_B$ be the event of traveling from A to H through waypoint B, and let $E_C$ be the event of traveling through C.  Then these are the composite events \medskip

$E_B= E_1\land ((E_3\land E_6)\lor(E_4\land E_7))\land E_9$ \medskip

$E_C = E_2\land E_5\land E_8\land E_9$, and \medskip

\noindent in accordance with the foregoing we can compute \medskip

${\rm Poss}(E_B)={\it min}({\rm Poss}(E_1),{\it max}({\it min}({\rm Poss}(E_3), {\rm Poss}(E_6)),{\it min}({\rm Poss}(E_4),$

\hskip 7.6em ${\rm Poss}(E_7)), {\rm Poss}(E_9)))$, \medskip

${\rm Poss}(E_C)={\it min}({\rm Poss}(E_2),{\rm Poss}(E_5),{\rm Poss}(E_8), {\rm Poss}(E_9))$, \medskip

\noindent and choose the path with the higher value.  If the path through waypoint B is chosen, then upon reaching that waypoint, consider $E_D=E_3\land E_6\land E_9$ and $E_E=E_4\land E_7\land E_9$, and compute \medskip

${\rm Poss}(E_D)={\it min}({\rm Poss}(E_3),{\rm Poss}(E_6),{\rm Poss}(E_9))$, and \medskip

${\rm Poss}(E_E)={\it min}({\rm Poss}(E_4),{\rm Poss}(E_7),{\rm Poss}(E_9))$  \medskip

\noindent and choose the path through D or E depending on which of these is higher.  In this manner one finds the path from A to H that is most possible to traverse at a speed that is at least as high as the designated speed limit.

As a point of terminology, one might refer to the process of traversing a leg of the journey as an {\it activity} and the completion of this process as an {\it event} indicated by the waypoint at the end of the leg.  In this case, the prerequisites and constraints apply to the activity, and the computed possibility measure represents the possibility of successfully completing this activity as well as the possibility of the actual occurrence of the event. 

\section{Comparison with Probability Theory}

Probability theory also provides a semantics for the logical $\lnot$, $\land$, and $\lor$.   The following borrows definitions and theorems from the textbook \cite{goldberg1986probability}.

The theory of probability is based on the concept of a {\it sample space} $S$ consisting of a set of {\it outcomes} $o$ from an experiment, like the tossing of one or more coins or the rolling of one or more six-sided dice.  An {event} $E$ is a subset of $S$. The event is said to {\it occur} as the result of an experiment if the outcome $o$ is a member of $E$.

Each outcome $o$ in $S$ is assumed to have been assigned a value, ${\rm Prob} (o)$, representing that outcome's probability of occurrence, with the properties (i) ${\rm Prob}(o)\ge 0$, for all $o$ in $S$, and (ii) the sum of the values ${\rm Prob}(o)$ over all the $o$ in $S$ equals 1.  Then, the probability ${\rm Prob}(E)$ of an event $E$ is computed as the sum of the probabilities of the outcomes in $E$.

Based on these definitions, it can be proven that, where $\cup$, $\cap$, and ${}^c$  represent ordinary set union, intersection, and complement, and $E$ and $F$ are any events, \medskip

${\rm Prob}(E\cup F)={\rm Prob}(E)+{\rm Prob}(F)-{\rm Prob}(E\cap F)$ \medskip

\noindent and \medskip

${\rm Prob}(E^c)=1-{\rm Prob}(E)$. \medskip

\noindent Two events $E$ and $F$ are said to be {\it independent} if \medskip

${\rm Prob}(E\cap F)={\rm Prob}(E)\times{\rm Prob}(F)$. \medskip 

An event $E$ can be associated with a logical proposition $p$ by taking $p$ as expressing ``$E$ has occurred'' and taking the truth value $v(p)$ as the probability value ${\rm Prob}(E)$.  Then, by virtue of the above, a logic may be defined for propositions representing independent events by setting \medskip

$v(\lnot p) = 1-v(p)$

$v(p\land q) =v(p)\times v(q)$

$v(p\lor q) =v(p)+v(q)-v(p\land q)$ \medskip

These definitions of the logical connectives might be considered for use in place of the multivalent interpretations throughout the foregoing discussion, ultimately, or ostensibly, providing a measure of the likelihood of successfully completing the journey along each of the possible paths, after which one could choose whichever plan is most likely to succeed. However, this probabilistic interpretation, in most applications, would produce unrealistic results.  This may be illustrated in terms of the foregoing waypoint navigation example.  For an event (leg traversal) $E_i$, the probability of this event would be computed as the product of the probabilities of the prerequisites and the opposites of the constraints.  Given that these would all be real numbers less than 1, this product would typically be very low, possibly close to 0.  Accordingly, this would imply that it is actually very unlikely that the leg traversal would occur.  This, however, does not square with common sense, and it therefore suggests that probability theory doesn't provide a realistic model of this kind or real-world situation.  

Last, it may be conjectured that the version of possibility presented in this paper provides a more intuitively compelling model of natural human reasoning than does probability theory for these kinds of applications.  The type of reasoning and decision making as illustrated in the case of Jane in Section 2 seems wholly natural and may well be a correct model of this kind of reasoning, i.e., using {\it min} and {\it max} for the logical {\it and} and {\it or}.   Of course, it would require suitable Cognitive Science experiments to verify that this is true.  But, in any case, it seems clear that humans do not take the trouble of computing the relevant probabilities for such logical combinations.  Day-to day human reasoning and decision making is much more fluid and intuitive and surely does not entail such complex cogitations. 

\section{Prospects for Applications}                                                    

Probability theory is concerned with prediction. In this context, one wants to know, given certain conditions, what is the likelihood that a given event will occur.  In contrast, possibility theory has no concern with prediction. Rather this addresses the situation where one wants to know, given certain preconditions, what is the possibility that a given event may occur.   Following the suggestion by Zadeh in \cite{zadeh1978fuzzy}, as discussed in the foregoing Section 2, the possibility value of an event may be interpreted as a measure of the {\it ease} with which that event may occur.  In an application such as the foregoing waypoint navigation example, an elementary event would the traversal of a leg of the journey.  Accordingly, the overall possibility measure of a path through the network from the start node, A, to the goal node, H, can be taken as representing the ease with which that path may be traversed, and the objective of the analysis is to determine which path is {\it easiest}.   Possibility values might also be taken as representing the {\it feasibility} of completing a task or the overall mission.  

These concepts can be applied generally to any planning application. The following suggests four further potential application domains.  \medskip

{\bf Robotics.} Robot planning involves finding a sequence of elementary actions that will take the robot from some start state to some goal state.  A well-known textbook on this subject is \cite{ghallab2004automated}. 

 A variety of methods for creating and executing plans have evolved.  The earliest one, STRIPS, cf. \cite{fikes1971strips} (also cf. \cite{nilsson1980principles, nilsson1998artificial}, describes the elementary actions that the robot may perform as logical if-then rules of the form ``if certain prerequisites are satisfied, then such-and-such action will be taken'', and the system uses an automated reasoning technique that attempts to prove that, given these rules, the goal can be achieved.  If a proof is found, then the steps of the proof comprise a viable plan.  

The success of STRIPS led to the creation of the Planning Domain Definition Language (PDDL) by Drew McDermott and others, cf. \cite{ghallab1998pddl}.  This provides a language for encoding the relevant domain knowledge and inference rules.  A recent treatment of this is the book by Patrik Haslum and others \cite{haslum2019introduction}.  

This planning methodology lends itself to application of the ideas being presented here in those situations where the prerequisites of the robot actions are satisfied with some degree of probability.   PDDL specifications also normally include some constraints, e.g., time of plan completion, which might also be represented in this theory.  This possibilistic information would be most relevant when there is more than one viable plan for reaching the goal, in which case one can select the plan that is most possible, i.e., easiest to complete.  

STRIPS/PDDL continues to be popular and has been employed in numerous robot competitions.  Other planning methodologies include hierarchical task networks, cf. \cite{bercher2019survey, georgievski2015htn}; behavior trees, cf. \cite{colledanchise2018behavior, iovino2022survey}, and synthesis methods, cf. \cite{kress2018synthesis}.  These may also be amenable to some version of this theory. \medskip

{\bf Organizational Project Planning.} The Critical Path Method (CPM) and the Program Evaluation and Review Technique (PERT), cf. \cite{armstrong1969critical, hansen1964practical, horowitz1967cpm, levin1966planning}, are project planning tools that are widely used throughout business, industry, and the US military.  CPM works with a graphical representation of the tasks that are required to complete a project, where some tasks must precede or follow other tasks, and some tasks may run in parallel with other tasks.  The graph thus comprises a network showing all the tasks that must be performed to reach from the start of the project to the end.  There are typically many paths through the network from the start to the end.  The {\it critical path} is the one that takes the longest time, thereby determining the minimum time required to completion.  PERT augments CPM by providing additional aspects for working with times, accommodating the fact that some tasks may complete early, and some may run late, accordingly dynamically adjusting the estimate of overall time while the work is in progress. 

If some or all of the tasks can be viewed as having prerequisites and/or constraints affecting the task completion, this would enable application of the current theory to determine which paths through the network are more feasible than others.  This could be useful additional information about the project plan inasmuch as the shortest-time path might not be the most feasible.  There would thus be a trade-off between time expectations and feasibility.  \medskip  

{\bf Computer Networks.}  Computer networks behave somewhat like the waypoint navigation problem described in Section 7.  Similarly with routing a vehicle from one place to another, computer networks send packets of information from one place to another.   Computer networks are much more complex than vehicular traffic systems, however.  The links between nodes can be of many different types, including both hard-wired---twisted copper, co-axial cable, ethernet---and wireless---Wi-Fi, radio frequency, microwave---where Wi-Fi can be 3G, 4G, 5G, etc., and where the latter has the issue that it only works over short distances.  

The present theory of possibility may be useful to provide further information about such networks. For any link in the network, the prerequisites might be very simple: only that the connection exists.  There can, however, be numerous potential constraints. These include congestion due to high traffic (as with vehicles on roadways), bandwidth constraints, noise, weather (electrical storms), etc. The research in this area is primarily concerned with the efficiency and reliability of transmission. This might be augmented with a measure of the feasibility of transmission.  \medskip

{\bf Computer Games.} The master's thesis by Braedon Warnick, cf.  \cite{warnick2022possibilistic} showed how this notion of possibility can be used to model the decision processes of non-player characters in a simple computer game.  A summary of this work had been published as \cite{warnick2022demonstration}.  Warnick agreed that this conforms to natural human decision processes, thereby providing the simulated actors with a more human-like behavior. 

\section{Conclusion}

This paper has proposed a simple but intuitively plausible procedure for computing the degree of possibility of an event insofar as the event may be viewed as having prerequisites that enable its occurrence and/or constraints that impede its occurrence.  The plausibility rests on the observation that the notion of possibility for this kind of an event is context dependent, where the context consists of those prerequisites and/or constraints.  The prerequisites may be satisfied, and the constraints may manifest, with specific numerical degrees of probability.  Thus, it seems reasonable to compute the possibility of the event in terms of these probabilities.  

This has the advantage that one can use the computational methods of statistical sampling to determine the indicated probabilities, and then use these probabilities to determine the event's possibility.  Thus, the overall method can be objective and computational.  This augments the currently existing subjective approaches and may serve as the foundation for future applications of possibility theory in practical real-world settings. 

Possibility theory can play an essential role in planning applications inasmuch as it provides a tool for evaluating the feasibility of alternative plans.  This opens potential opportunities to employ possibility theory in areas such as robot mission planning, organizational planning, computer networks, and computer games.

\begin{acks}
Lotfi A. Zadeh was the external examiner on my doctoral dissertation.  He was my advisor's advisor.  When I entered the Systems Science PhD Program at Portland State University in 1975, I met George L. Lendaris, who had been a student of Lotfi at Berkeley.  George saw that I had completed a master's thesis in mathematical logic and showed me some of Lotfi's papers.  I immediately became interested in the prospect of formalizing human reasoning with imprecise information, and this ultimately became the topic of my doctoral research.  George connected me with Lotfi, and I went to Berkeley once to give a talk.  Lotfi helped me in many ways over the years, including with obtaining my current job at Florida State University and with funding for my research.  He was an extraordinarily brilliant, creative, and wise human being, with a kind heart and a generous soul.  He should have received the Turing award, but he wasn't interested.  He said that his legacy would be his publications.  Lotfi was one of the giants.      
\end{acks}

\printbibliography

\appendix
\section{Appendix}

A software package had been produced by some undergraduate students at Florida State University that can take a street network of arbitrary complexity as input and output a listing showing a most possible solution path together with a graphical visualization. The most recent student to work on this is Analia Castellanos, who extended a system that had been developed previously by a team consisting of Maia Huber, Samuel Cohen, and Julian Olivier. An example input file is shown here.  This system, together with three sample input files and user instructions can be obtained at https://github.com/danielgschwartz/VehicleWaypointNavigation.

A sample input file is as follows.  This uses a slightly more complex street network than in Figure 1.  Here a leg of the network is referred to as a step, and each step is described by its start and end nodes, its prerequisites and constraints, and, for each prerequisite and constraint, an assumed probability of occurrence. As in the foregoing example, it is here assumed for simplicity of the illustration that all legs have the same prerequisites and constraints, but in general it is allowed that these be different for different legs.                                                   

\begin{verbatim}
Number of steps: 10
Origin: A
Destination: H

Step 1 origin: A
Step 1 destination: B
Constraint(s): red light, car accident
Prerequisite(s): car turns on, gas in tank
Probability of "red light": 0.2
Probability of "car accident": 0.2
Probability of "car turns on": 0.9
Probability of "gas in tank": 0.8

Step 2 origin: A
Step 2 destination: C
Constraint(s): road closure, traffic
Prerequisite(s): gas in tank
Probability of "road closure": 0.1
Probability of "traffic": 0.1
Probability of "gas in tank": 0.9

Step 3 origin: B
Step 3 destination: D
Constraint(s): traffic, car accident
Prerequisite(s): gas in tank
Probability of "traffic": 0.4
Probability of "car accident": 0.2
Probability of "gas in tank": 0.8

Step 4 origin: B
Step 4 destination: E
Constraint(s): red light, detour
Prerequisite(s): car turns on, gas in tank
Probability of "red light": 0.3
Probability of "detour": 0.1
Probability of "car turns on": 0.9
Probability of "gas in tank": 0.8

Step 5 origin: C
Step 5 destination: F
Constraint(s): car accident, road closure
Prerequisite(s): gas in tank
Probability of "car accident": 0.2
Probability of "road closure": 0.3
Probability of "gas in tank": 0.9

Step 6 origin: D
Step 6 destination: G
Constraint(s): traffic, construction
Prerequisite(s): gas in tank
Probability of "traffic": 0.4
Probability of "construction": 0.3
Probability of "gas in tank": 0.8

Step 7 origin: E
Step 7 destination: G
Constraint(s): car accident, traffic
Prerequisite(s): gas in tank
Probability of "car accident": 0.3
Probability of "traffic": 0.4
Probability of "gas in tank": 0.9

Step 8 origin: F
Step 8 destination: G
Constraint(s): detour, construction
Prerequisite(s): gas in tank
Probability of "detour": 0.5
Probability of "construction": 0.3
Probability of "gas in tank": 0.7

Step 9 origin: G
Step 9 destination: H
Constraint(s): traffic, road closure
Prerequisite(s): gas in tank
Probability of "traffic": 0.3
Probability of "road closure": 0.2
Probability of "gas in tank": 0.9

Step 10 origin: C
Step 10 destination: G
Constraint(s): detour, construction
Prerequisite(s): gas in tank
Probability of "detour": 0.1
Probability of "construction": 0.1
Probability of "gas in tank": 0.9

Desired path origin: A
Desired path destination: H
\end{verbatim}

The output listing generated by this input file is as follows, and the resulting graphical visualization is shown in Figure 2. Here the start node is at the top and the goal node is at the bottom.  The numbers on the legs are the computed possibility degrees, and the solution path having the highest possibility is shown in green. 

\begin{verbatim}
max prereqs: 2
max constraints: 2
first segment row: ['A', 'B', 0.9, 0.8, 0.2, 0.2]

Segments created:
['A', 'B', 0.9, 0.8, 0.2, 0.2]
['A', 'C', 0.9, 1.0, 0.1, 0.1]
['B', 'D', 0.8, 1.0, 0.4, 0.2]
['B', 'E', 0.9, 0.8, 0.3, 0.1]
['C', 'F', 0.9, 1.0, 0.2, 0.3]
['D', 'G', 0.8, 1.0, 0.4, 0.3]
['E', 'G', 0.9, 1.0, 0.3, 0.4]
['F', 'G', 0.7, 1.0, 0.5, 0.3]
['G', 'H', 0.9, 1.0, 0.3, 0.2]
['C', 'G', 0.9, 1.0, 0.1, 0.1]
Segment A-B: Poss = 0.8
Segment A-C: Poss = 0.9
Segment B-D: Poss = 0.6
Segment B-E: Poss = 0.7
Segment C-F: Poss = 0.7
Segment D-G: Poss = 0.6
Segment E-G: Poss = 0.6
Segment F-G: Poss = 0.5
Segment G-H: Poss = 0.7
Segment C-G: Poss = 0.9

Graph adjacency list:
A -> ['B', 'C']
B -> ['D', 'E']
C -> ['F', 'G']
D -> ['G']
E -> ['G']
F -> ['G']
G -> ['H']
Possibility of Path A-B-D-G-H: 0.6
Possibility of Path A-B-E-G-H: 0.6
Possibility of Path A-C-F-G-H: 0.5
Possibility of Path A-C-G-H: 0.7
Choosing Path A-C-G-H with Possibility 0.7
Possibility of Path C-F-G-H: 0.5
Possibility of Path C-G-H: 0.7
Possibility of Path G-H: 0.7
Choosing Path C-G-H with Possibility 0.7
Full chosen path: A-C-G-H
Saved: final_path.png
\end{verbatim}

\begin{figure}[h]
\centerline{\includegraphics[width=4in]{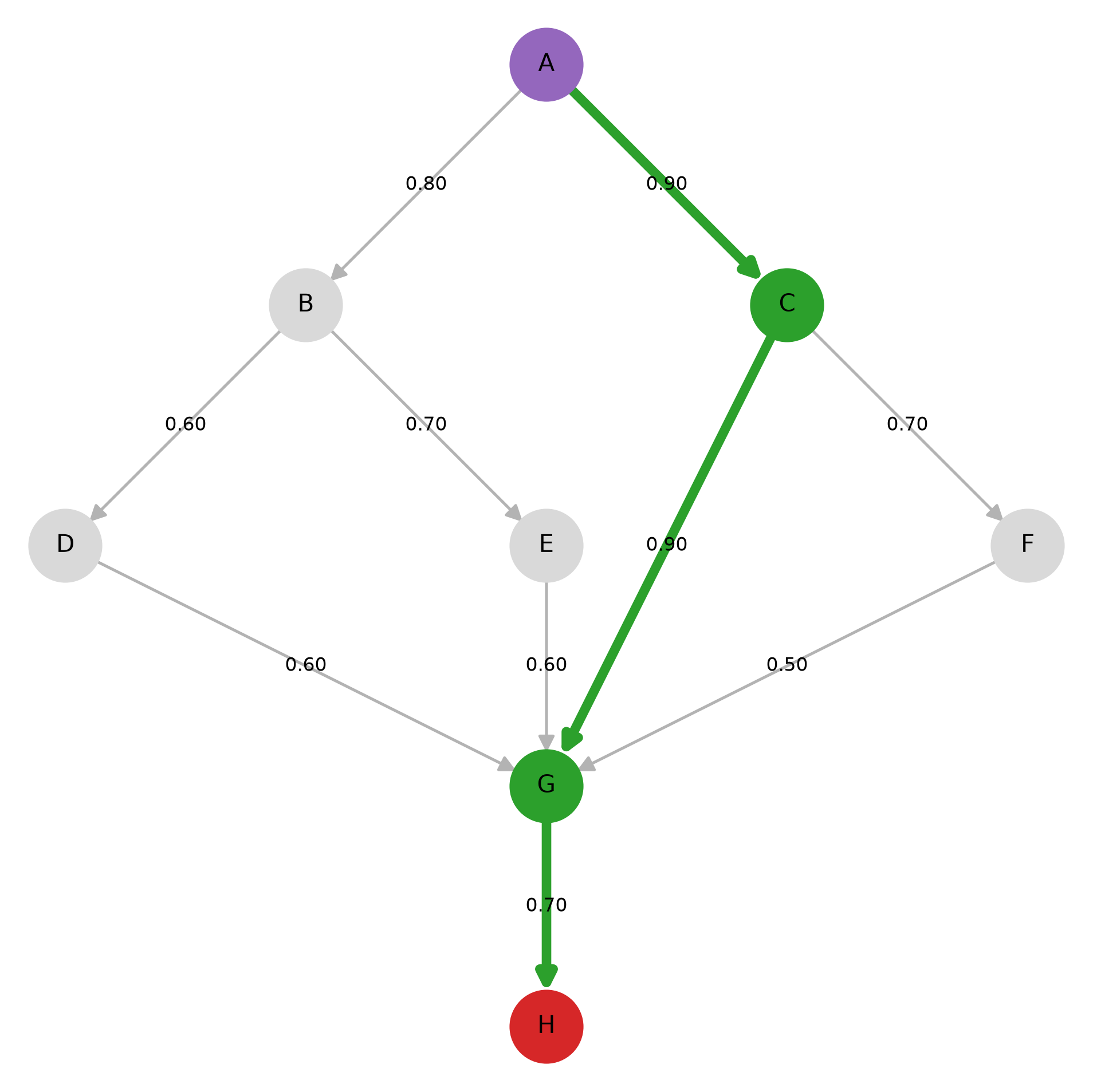}}
\caption{Street network with solution path.}
\end{figure}

\end{document}